\documentclass[a4paper,fleqn]{cas-dc}

\usepackage[authoryear,longnamesfirst]{natbib}
\usepackage{algorithm}
\usepackage{algpseudocode}
\usepackage{booktabs} 
\usepackage{tabularx}
\usepackage{array}
\usepackage{hyperref}
\usepackage{etoolbox}
\def\tsc#1{\csdef{#1}{\textsc{\lowercase{#1}}\xspace}}
\tsc{WGM}
\tsc{QE}
\tsc{EP}
\tsc{PMS}
\tsc{BEC}
\tsc{DE}

\AtBeginEnvironment{figure}{\small}
\AtBeginEnvironment{figure*}{\small}
\AtBeginEnvironment{table}{\small}
\AtBeginEnvironment{table*}{\small}
\begin{document}
\let\WriteBookmarks\relax
\def\floatpagepagefraction{1}
\def\textpagefraction{.001}

\shorttitle{Constraint-Aware Hierarchical Search for Regulation-Driven Fine-Grained Classification}

\shortauthors{Siyu Wang et~al.}

\title [mode = title]{Constraint-Aware Hierarchical Search for Regulation-Driven Fine-Grained Classification}                      



%
\author[1,2]{Siyu Wang}[type=editor]
\fnmark[1]


\ead{wangsiyu2022@gusulab.ac.cn}



\affiliation[1]{organization={Gusu Laboratory of Materials},
    city={Suzhou},
    country={China}}

\affiliation[2]{organization={Chongqing Institute of Engineering, Chongqing Engineering Research Center for Intelligent Applications of Financial Big Data},
    city={Chongqing},
    country={China}}

\author[3]{Wei Tan}[style=chinese]
\fnmark[1]

\ead{tanwei@dces.cn}
\affiliation[3]{organization={Suzhou Digital China Wuxin Intelligent Technology Co., Ltd.},
    city={Suzhou},
    country={China}}

\author[1]{Lulu Chen}[style=chinese]
\cormark[1]

\cortext[cor1]{Corresponding author}

\ead{chenlulu2021@gusulab.ac.cn}
\fntext[fn1]{Siyu Wang and Wei Tan contributed equally to this work and should be considered co-first authors.}


\begin{abstract}
Tasks such as customs tariff classification, export control categorization, and standards-based equipment coding require assigning an input instance to a fine-grained class under an explicit regulatory hierarchy. Unlike standard text classification, the correct label in these tasks is not determined by semantic similarity alone, but by rule-defined boundaries, threshold conditions, exclusion clauses, definitions, and local exceptions. As a result, two highly similar inputs may require different labels, while a retrieved passage that appears relevant may still be inapplicable under the governing rules. Existing flat classifiers, hierarchical text classification methods, and retrieval-augmented LLM systems are not designed to jointly enforce hierarchical validity, rule consistency, and fine-grained boundary reasoning. In this paper, we formulate this setting as \textbf{regulation-driven fine-grained hierarchical classification}, where an external instance must be assigned to a fine-grained class through a valid path in a regulatory hierarchy and supported by auditable evidence. We construct four benchmark datasets from representative regulation-intensive scenarios and validate the annotations through an expert-in-the-loop process. We further propose a \textbf{constraint-aware hierarchical search} framework that converts regulatory documents into a searchable tree, retrieves only valid local candidate nodes, and uses structured regulatory fields with evidence snippets to guide each next-hop decision. Experiments show that our method achieves the best mean accuracy on all four datasets and provides interpretable decision paths, with the largest gains on cases involving fine-grained neighboring categories and rule-based boundary conditions.
\end{abstract}




\maketitle

\section{Introduction}

\begin{figure*}
    \centering
    \includegraphics[width=0.9\linewidth]{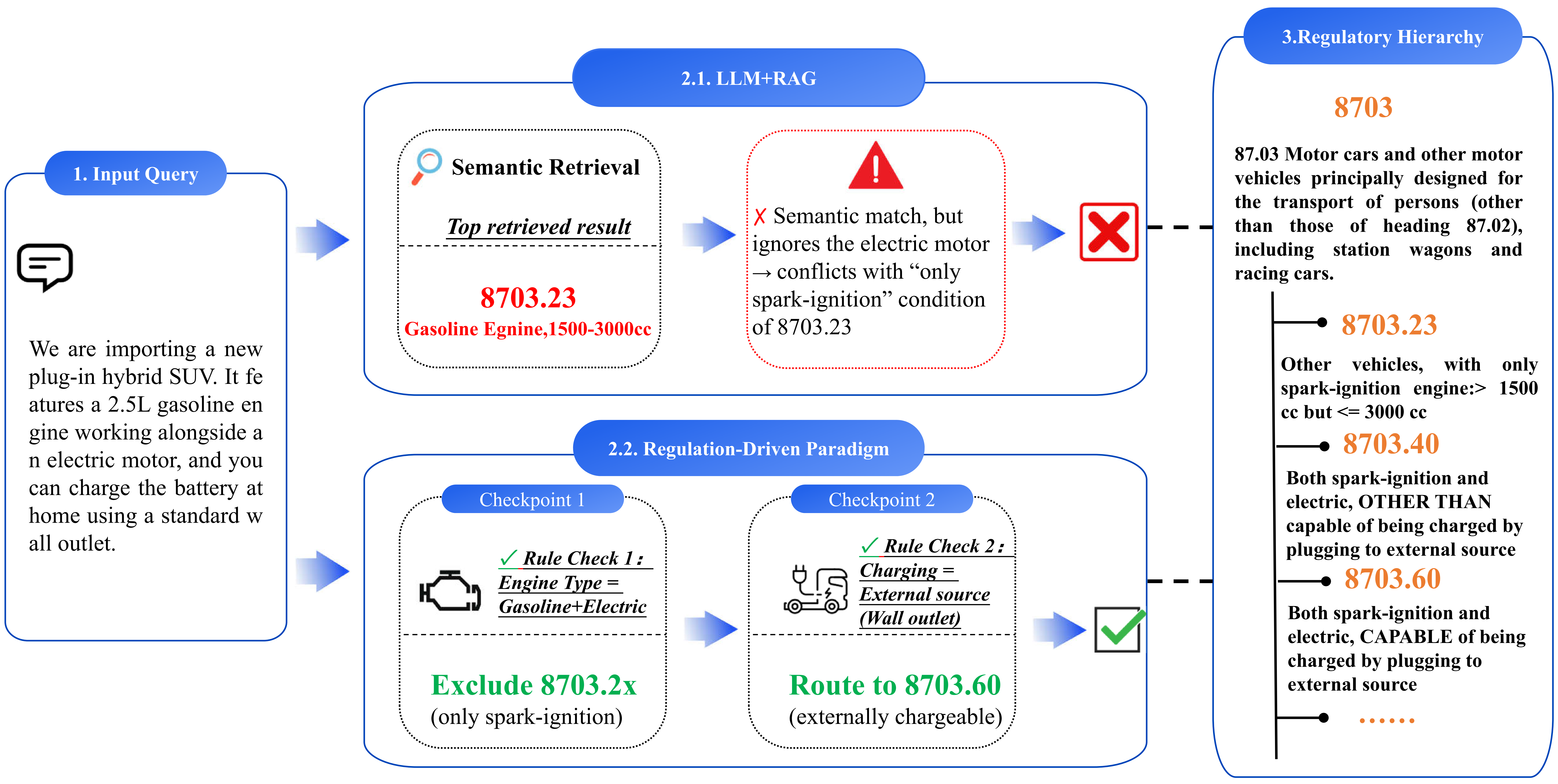}
    \caption{A motivating example from customs tariff classification. A semantic match may retrieve a plausible but invalid code, while a regulation-driven decision must follow the hierarchy and check the decisive conditions.}
    \label{fig:example}
\end{figure*}

Many real-world classification problems look deceptively similar to ordinary text classification: given a description of an object, case, or request, a system must assign it to the correct category. In institutional and regulatory settings, however, the category boundary is defined by an external rule system rather than by semantic similarity alone. Such systems often contain structured taxonomies, inclusion criteria, threshold conditions, exclusion clauses, definitions, and local exceptions. A single attribute can redirect the decision to a different branch, and a semantically relevant rule can still be inapplicable if it belongs to the wrong regulatory scope.

Recent advances in large language models (LLMs), dense retrieval, and retrieval-augmented generation (RAG) have achieved strong performance on general text classification, question answering, and other knowledge-intensive tasks \cite{karpukhin2020dpr,lewis2020rag,guu2020realm,gao2023rag,du2026rag}. However, regulatory classification requires a decision to satisfy the hierarchy, the governing rules, and the fine-grained boundary conditions at the same time. A flat classifier may predict a label without checking whether it is reached through a valid branch. A retriever may return passages that are semantically related but legally out of scope. An LLM may generate a plausible code, but without sufficient domain-specific regulatory knowledge, it may fail to correctly interpret the local conditions, exclusions, and thresholds that distinguish neighboring categories. The challenge is therefore not only to find relevant labels or passages, but also to make a stable and auditable decision under explicit hierarchical and regulatory constraints.

The classification of customs tariffs under the Harmonized System (HS) provides a concrete example. In real-world customs declaration, company staff often determine the correct HS code from a product description by consulting tariff schedules, explanatory notes, and related regulatory provisions. This process is time-consuming, knowledge-intensive, and sensitive to fine-grained product attributes, because a small difference in wording may lead to a different code and different legal consequences. AI assistance is therefore valuable, but reliable automation requires more than semantic matching.

As shown in Figure \ref{fig:example}, consider a product description: \textit{"a plug-in hybrid SUV with a 2.5L gasoline engine, an electric motor, and a battery that can be charged from a standard wall outlet."} A semantic system may focus on \textit{"2.5L gasoline engine"} and predict HS code 8703.23, which covers vehicles with only spark-ignition engines. It may also recognize the hybrid signal but predict HS code 8703.40, which covers hybrid vehicles that are not externally chargeable. Both predictions are plausible by surface similarity, but both are invalid under the HS rules. The electric motor excludes the \textit{"only spark-ignition"} branch, and the wall-outlet charging condition routes the item to the externally chargeable branch. The correct decision therefore requires following the HS hierarchy, rejecting invalid neighboring codes, and selecting HS code 8703.60. This example shows why the problem is not simply label matching: the system must map natural language attributes to rule conditions, navigate a hierarchy, compare fine-grained boundaries, and preserve auditable evidence.

This pattern of decisions appears in many domains, including customs tariff classification, export control categorization, hazardous goods transportation classification, medical coding, and chemical compliance classification. Although these applications differ in content, they share the same underlying mechanism: the label space is hierarchical, the decision procedure is constrained by explicit rules, and neighboring categories are separated by small but decisive attributes. Existing hierarchical text classification methods model label dependencies and hierarchical consistency \cite{silla2011survey,wehrmann2018hierarchical,zhou2020hierarchy,wang2022hierarchical}, while fine-grained classification focuses on distinguishing closely related categories \cite{ling2012fine,ren2016afet,choi2018ultra}. Legal and regulatory NLP further studies legal-domain representation learning, document classification, legal reasoning benchmarks, and compliance-oriented analysis \cite{chalkidis2020legalbert,zhong2020legalai,chalkidis2022lexglue,guha2023legalbench,regbr2022}. These lines of work are highly relevant, but they do not fully capture the setting where an external instance must be assigned to a fine-grained class by jointly satisfying a regulatory hierarchy, explicit rules, and boundary conditions.

We therefore argue that this setting should not be treated merely as a special case of flat text classification, conventional hierarchical classification, or semantic retrieval. Instead, we formulate it as a unified task paradigm, which we call \textbf{Regulation-driven Fine-grained Hierarchical Classification}. In this task, the system must not only predict a final category, but also produce a hierarchy-consistent decision path and evidence-grounded rationale that explain why the selected class is valid and why close alternatives are rejected.

To address this paradigm, we propose a \textbf{Constraint-aware Hierarchical Search Framework}. The framework first organizes regulatory documents into a searchable tree with node-level and evidence-level indexes. At inference time, it avoids one-shot prediction over a flat label space and instead performs local search: at each step, it retrieves candidate child nodes within the current regulatory scope, constructs evidence-enriched candidate packages, and lets an LLM choose the next hop by comparing both semantic relevance and explicit regulatory conditions. This design turns classification into a controlled traversal process, where each decision remains within the valid hierarchy and can be supported by retrieved evidence.

In summary, our main contributions are as follows.

1. To the best of our knowledge, this work is the first to formulate \textbf{Regulation-driven Fine-grained Hierarchical Classification} as a unified task paradigm, rather than treating it as a special case of flat text classification, conventional hierarchical classification, or pure semantic retrieval.

2. We propose a \textbf{Constraint-aware Hierarchical Search Framework} that combines offline construction of a searchable regulatory tree and evidence index with online local decision making under hierarchical and regulatory constraints. The framework further aggregates evidence along the selected path for final decision verification and can flag low-confidence cases when regulatory support is insufficient.

3. We construct four benchmark datasets from representative regulation-intensive classification scenarios and conduct extensive experiments against LLM and retrieval-augmented baselines. The results show that the proposed framework improves final-label accuracy while providing interpretable decision paths and evidence-grounded outputs, especially in cases involving fine-grained neighboring categories.

\section{Related Works}
Our work is related to hierarchical text classification, fine-grained classification, regulatory and legal NLP, and retrieval-augmented decision systems. These areas provide useful building blocks, but none directly studies the setting where an external instance must be assigned to a fine-grained regulatory class by following a rule-defined hierarchy.

\subsection{Hierarchical Text Classification}

Hierarchical text classification assigns instances to labels organized in a tree or directed acyclic graph. Classical surveys summarize strategies such as flat classification, local classifiers, and global hierarchical models \cite{silla2011survey}, while neural methods model label dependencies with hierarchical networks, structure-aware encoders, or contrastive representations \cite{kowsari2017hdltex,wehrmann2018hierarchical,zhou2020hierarchy,wang2022hierarchical,ji2024domain,karl2025hydra}. These methods are designed to improve taxonomy-consistent label prediction. In contrast, our task treats the hierarchy as a regulatory decision space: each transition must remain valid under parent-child constraints, and local rules such as thresholds, exclusions, and definitions may invalidate semantically similar candidates.

\subsection{Fine-Grained Classification}

Fine-grained classification focuses on distinguishing closely related categories with subtle attribute differences. Representative NLP tasks include fine-grained and ultra-fine entity typing \cite{ling2012fine,ren2016afet,choi2018ultra,mtumbuka2024encore,komarlu2024ontotype}, as well as fine-grained legal claim classification where hierarchical class structure can improve prediction \cite{dayanik2021using}. Our task shares this boundary-sensitive nature, but the source of the boundary is different. Conventional fine-grained classification usually learns distinctions from labeled examples or semantic representations, whereas regulation-driven classification must follow externally specified rules and provide evidence-grounded justification for both the selected class and rejected neighboring classes.

\subsection{Regulatory and Legal NLP}

Legal and regulatory NLP studies legal document classification, compliance analysis, contract obligation detection, legal reasoning, and regulatory information extraction. Domain-specific models and benchmarks such as LEGAL-BERT, LexGLUE, CUAD, LegalBench, and RegBR have advanced representation learning and evaluation in legal or regulatory settings \cite{chalkidis2020legalbert,zhong2020legalai,chalkidis2022lexglue,hendrycks2021cuad,guha2023legalbench,regbr2022,marino2025aireg,asai2024self,zhu2024atm,joshi2024tur}. These works often classify legal texts themselves, extract obligations, predict legal outcomes, or assess compliance states. By contrast, our task uses regulatory texts as the decision authority for assigning an external object or case to the most specific applicable class in a regulatory hierarchy.

\subsection{Retrieval-Augmented and LLM-Based Decision Systems}

Dense retrieval and retrieval-augmented generation provide strong foundations for knowledge-intensive NLP \cite{karpukhin2020dpr,guu2020realm,lewis2020rag,izacard2021fid,gao2023rag,huang2024survey}. Recent RAG systems further incorporate graph structures, hierarchical indexing, or agentic reasoning, including GraphRAG, LightRAG, LinearRAG, MA-RAG, and ReAct-style reasoning \cite{edge2024graphrag,guo2024lightrag,zhuang2025linearrag,nguyen2025marag,yao2023react,zhu2025knowledge,huang2025retrieval,jin2025hierarchical}. These systems retrieve semantically relevant evidence and use LLMs to generate answers or explanations. However, semantic relevance alone is insufficient in regulation-driven fine-grained hierarchical classification: a passage may be relevant but inapplicable under the current branch, or a plausible label may violate a local exclusion or threshold condition. Our method therefore restricts retrieval to valid local candidates, attaches rule-grounded evidence to candidate nodes, and performs stepwise next-hop decisions within the regulatory hierarchy.

\section{Constraint-Aware Hierarchical Search Framework}

\begin{figure*}
    \centering
    \includegraphics[width=0.7\linewidth]{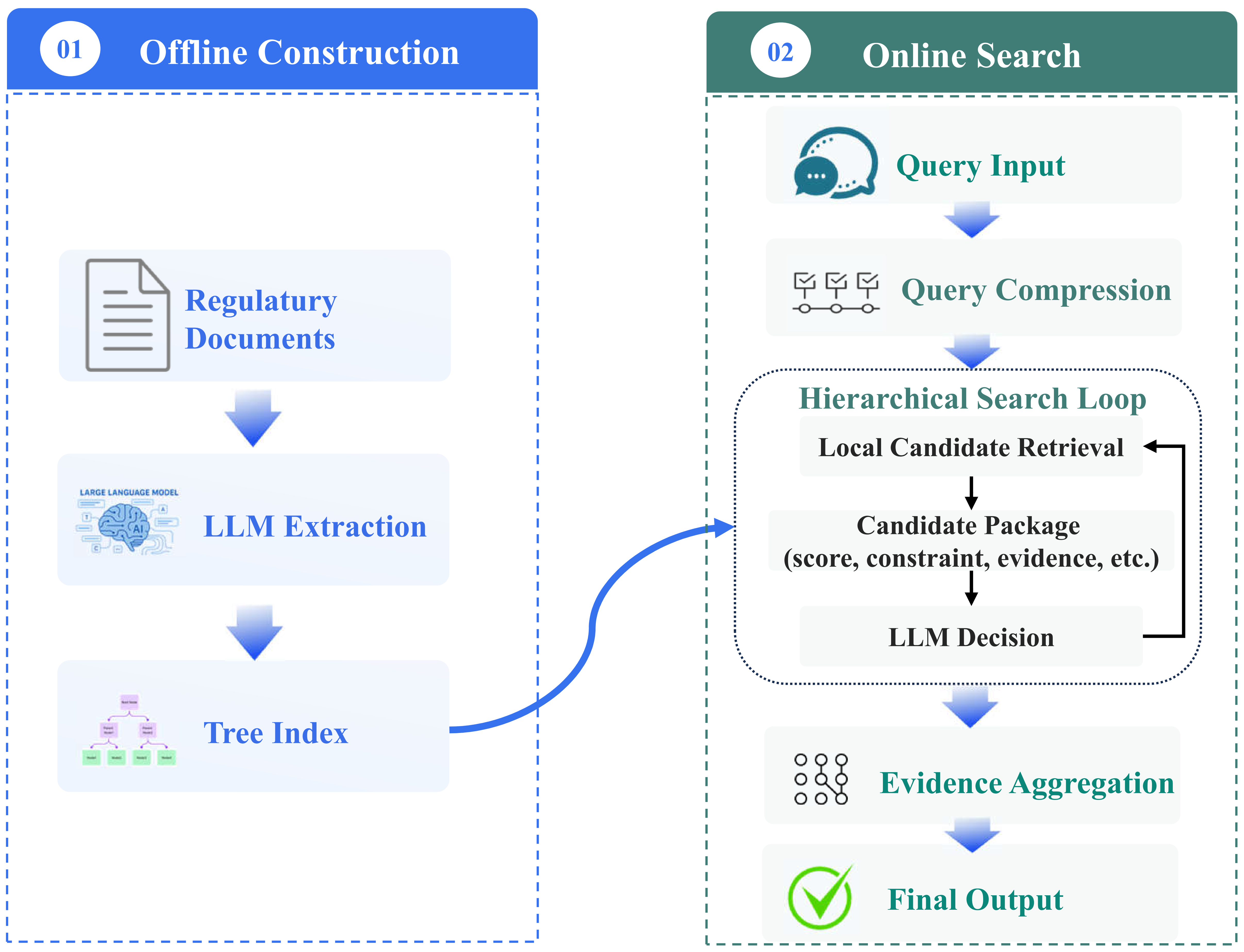}
    \caption{Overview of the proposed framework. Offline, regulatory documents are converted into a searchable tree with node and evidence indexes. Online, the model searches the tree step by step, compares only valid local candidates, and aggregates evidence along the selected path.}
    \label{fig:model}
\end{figure*}

Our method has two stages, as shown in Figure \ref{fig:model}. Offline, regulatory documents are parsed into a hierarchical tree and indexed at both the node and evidence levels. Online, the system performs constraint-aware hierarchical search, where each local decision uses both semantic relevance and explicit regulatory conditions.

\subsection{Problem Definition}

We study \textbf{regulation-driven fine-grained hierarchical classification}, where an input instance must be assigned to a fine-grained class under a regulatory taxonomy. Unlike standard text classification, the target space is hierarchical, and the final decision must satisfy applicable regulatory rules. Formally, we represent the regulatory knowledge space as a heterogeneous hierarchical tree

\begin{equation}
\mathcal{T} = (\mathcal{V}, \mathcal{A}, v_{\mathrm{root}}),
\end{equation}
where $\mathcal{V}$ is the set of nodes, $\mathcal{A}$ is the set of parent-child edges, and $v_{\mathrm{root}} \in \mathcal{V}$ is the root node. A node may represent either a final class or an intermediate regulatory unit, such as a document, section, chapter, heading, subheading, explanatory note, or reference node. For each node $v \in \mathcal{V}$, we define

\begin{equation}
v = (\mathrm{id}_v, \mathrm{code}_v, \mathrm{title}_v, \mathrm{type}_v, \mathrm{text}_v, \mathcal{N}_v, \mathcal{X}_v, \mathcal{D}_v, \mathcal{Z}_v, \mathrm{Child}(v)),
\end{equation}
where $\mathrm{id}_v$, $\mathrm{code}_v$, $\mathrm{title}_v$, $\mathrm{type}_v$, and $\mathrm{text}_v$ denote the node identifier, regulatory code, title, type, and original text, respectively. The sets $\mathcal{N}_v$, $\mathcal{X}_v$, and $\mathcal{D}_v$ denote general notes, exclusion clauses, and definitional provisions, and $\mathcal{Z}_v$ contains associated evidence snippets. The child set of $v$ is defined as

\begin{equation}
\mathrm{Child}(v) = \{u \mid (v, u) \in \mathcal{A}\}.
\end{equation}

Given an input instance $x$, such as a natural language product description, the system predicts a decision path from the root to a terminal or decision node:

\begin{equation}
p(x) = (v_0, v_1, \ldots, v_L), \quad v_0 = v_{\mathrm{root}},
\end{equation}
where each transition follows the tree structure:

\begin{equation}
v_{t+1} \in \mathrm{Child}(v_t), \quad t = 0, 1, \ldots, L-1.
\end{equation}

The final node $v_L$ denotes the predicted fine-grained class. The output consists of

\begin{equation}
y(x) = (\hat{c}, p(x), \mathcal{S}(x), \eta),
\end{equation}
where $\hat{c}$ is the predicted regulatory class, $p(x)$ is the decision path, $\mathcal{S}(x)$ is the supporting evidence selected from relevant nodes and snippets, and $\eta$ is a natural-language rationale. This formulation casts the task as constrained search over a hierarchical regulatory space.


\subsection{Offline Tree and Index Construction}

Before online classification, the raw regulatory documents are converted into a tree that can be searched locally. The goal of this stage is simple: preserve the regulatory hierarchy, keep rule-bearing text attached to the nodes where it applies, and prepare retrieval indexes for later branch selection. This avoids treating a long regulation as an unstructured text pool, where headings, category descriptions, notes, definitions, and exclusions may be mixed together.

Given a collection of regulatory documents, we first split the raw text into locally coherent blocks

\begin{equation}
    \mathcal{B} = \{b_1, b_2, \ldots, b_m\}.
\end{equation}

The segmentation uses document cues such as headings, numbering patterns, paragraph boundaries, and list markers. Each block $b_i$ is intended to contain a complete regulatory unit or a small group of related units, while metadata such as document title, source file, and page range is kept for traceability.

Second, an LLM-based extractor converts each block into structured regulatory units. Each unit follows the node representation in Section 3.1, including its code, title, type, original text, notes $\mathcal{N}_v$, exclusions $\mathcal{X}_v$, definitions $\mathcal{D}_v$, evidence snippets $\mathcal{Z}_v$, and extracted children $\mathrm{Child}(v)$. The key point is that evidence snippets do not determine the hierarchy by themselves. They preserve the original wording of relevant provisions, while the parent-child relation is taken from the extracted \texttt{children} structure and later materialized into tree edges.

Third, the extracted units are normalized, deduplicated, and recursively inserted into the global tree. If an extracted node $v$ contains child units, each child $u \in \mathrm{Child}(v)$ is added to the node set and linked by a parent-child edge:

\begin{equation}
\mathcal{V} \leftarrow \mathcal{V} \cup \{u\}, \quad
\mathcal{A} \leftarrow \mathcal{A} \cup \{(v,u)\}.
\end{equation}

Applying this operation from the root downward produces the searchable regulatory tree $\mathcal{T} = (\mathcal{V}, \mathcal{A}, v_{\mathrm{root}})$. More implementation details and an illustrative construction example are provided in Appendix~\ref{app:ours}.

Finally, we build two retrieval indexes. For each node $v$, we concatenate its title, original text, notes, exclusions, and definitions into a node retrieval text and encode it using an embedding model $\phi(\cdot)$:

\begin{equation}
\mathbf{h}_v = \phi(v).
\end{equation}

For each evidence snippet $z \in \mathcal{Z}_v$, we encode its text together with source context:

\begin{equation}
\mathbf{h}_z = \phi(z).
\end{equation}

The node index supports locating plausible child nodes during hierarchical search, while the evidence index supports rule verification and rationale generation. Together, the tree and the two indexes provide the offline representation used by the online constraint-aware search procedure.

\subsection{Online Constraint-Aware Hierarchical Search}

Algorithm~\ref{alg:online_search} summarizes the online constraint-aware hierarchical search procedure. Given an input instance, the system first compresses the raw query into a retrieval-oriented form that keeps the attributes needed for classification. The search then starts from the root node and moves downward. At each step, $\operatorname{TopM}$ retrieves plausible child nodes under the current node, and $\operatorname{TopE}$ retrieves supporting evidence within each candidate node. The system builds a candidate package with both semantic and regulatory information, and the decision model selects the next hop or stops. After the path is fixed, evidence from the visited nodes is aggregated for final verification and rationale generation.

\begin{algorithm}[t]
\caption{Online Constraint-Aware Hierarchical Search}
\label{alg:online_search}
\begin{algorithmic}[1]
\Require Input instance $x$, regulatory tree $\mathcal{T}$, node embeddings $\{\mathbf{h}_v\}$, evidence embeddings $\{\mathbf{h}_z\}$
\Ensure Final decision $\hat{y} = (\hat{c}, p(x), \mathcal{S}(x), \eta)$

\State $q^{\mathrm{ret}} \gets \Psi(x)$
\State $p(x) \gets (v_{\mathrm{root}})$
\State $v_t \gets v_{\mathrm{root}}$

\While{$\mathrm{Child}(v_t) \neq \emptyset$ and search does not stop}
    \ForAll{$u \in \mathrm{Child}(v_t)$}
        \State $s_{u \mid v_t, x} \gets \cos(\phi(q^{\mathrm{ret}}), \mathbf{h}_u)$
    \EndFor
    \State $\mathcal{C}_t(x) \gets \operatorname{TopM}_{u \in \mathrm{Child}(v_t)} s_{u \mid v_t, x}$

    \ForAll{$u \in \mathcal{C}_t(x)$}
        \State $\mathcal{S}(u \mid x) \gets \operatorname{TopE}_{z \in \mathcal{Z}_u}
        \cos(\phi(q^{\mathrm{ret}}), \mathbf{h}_z)$
        \State Construct candidate package $\Gamma(u)$
    \EndFor

    \State $(v_{t+1}, \sigma_t, \rho_t, \eta_t) \gets
    F_\theta(x, v_t, \{\Gamma(u)\}_{u \in \mathcal{C}_t(x)})$

    \If{$\sigma_t = 1$}
        \State \textbf{break}
    \EndIf

    \State Append $v_{t+1}$ to $p(x)$
    \State $v_t \gets v_{t+1}$
\EndWhile

\State $\mathcal{S}(x) \gets \operatorname{MergeTop}_{v \in p(x)} \mathcal{S}(v \mid x)$
\State $\hat{y} \gets G_\theta(x, p(x), \mathcal{S}(x))$
\State \Return $\hat{y}$
\end{algorithmic}
\end{algorithm}

\subsubsection{Query Compression}

The raw input $x$ may contain redundant or informal details. Before retrieval, an LLM compresses it into a retrieval-oriented query by keeping key attributes such as product name, material, function, usage, process, power source, quantities, and limiting conditions:

\begin{equation}
q^{\mathrm{ret}} = \Psi(x),
\end{equation}
where $\Psi(\cdot)$ is the query compression function. The compressed query $q^{\mathrm{ret}}$ reduces retrieval noise while preserving attributes needed for hierarchical localization.

\subsubsection{Local Candidate Retrieval}

At depth $t$, let the current node be $v_t$. The system retrieves candidates only from its direct children $\mathrm{Child}(v_t)$. It does not compare the input with the whole tree at this step. For each child node $u$, the local similarity score is

\begin{equation}
s_{u \mid v_t, x}
= \cos\big(\phi(q^{\mathrm{ret}}), \mathbf{h}_u\big),
\quad u \in \mathrm{Child}(v_t).
\end{equation}

The top-$M$ children are kept as local candidate next hops:

\begin{equation}
\mathcal{C}_t(x)
= \operatorname{TopM}_{u \in \mathrm{Child}(v_t)}
s_{u \mid v_t, x}.
\end{equation}

These candidates are possible next nodes, not complete paths. This local retrieval step keeps the search inside the valid regulatory scope and avoids irrelevant labels from other branches.

\subsubsection{Candidate Package Construction}

For each candidate child node $u \in \mathcal{C}_t(x)$, the system builds a package containing its identity, retrieval score, local rule context, and supporting evidence:

\begin{equation}
\Gamma(u) =
(\mathrm{id}_u, \mathrm{code}_u, \mathrm{title}_u, \mathrm{type}_u,\mathrm{text}_u,
s_u, \mathcal{N}_u, \mathcal{X}_u, \mathcal{D}_u, \mathcal{S}(u \mid x)).
\end{equation}

Here, $s_u$ denotes $s_{u \mid v_t, x}$. The evidence set $\mathcal{S}(u \mid x)$ contains the most relevant snippets retrieved from $\mathcal{Z}_u$:

\begin{equation}
\mathcal{S}(u \mid x)
= \operatorname{TopE}_{z \in \mathcal{Z}_u}
\cos\big(\phi(q^{\mathrm{ret}}), \mathbf{h}_z\big).
\end{equation}

This package lets the decision model check notes, exclusions, definitions, and evidence, rather than relying on semantic similarity alone.

\subsubsection{Next-Hop Decision}

Given the input, current node, and candidate packages, an LLM-based local decision function $F_\theta$ selects the next hop or stops:

\begin{equation}
(v_{t+1}, \sigma_t, \rho_t, \eta_t)
= F_\theta\big(x, v_t, \{\Gamma(u)\}_{u \in \mathcal{C}_t(x)}\big).
\end{equation}

Here, $v_{t+1}$ is the selected next hop, $\sigma_t$ is the stop signal, $\rho_t$ is the local confidence score, and $\eta_t$ is the rationale. The model checks both semantic relevance and regulatory constraints. Therefore, a semantically similar child can still be rejected if its exclusions, thresholds, or conditions conflict with the input.

The search stops when any of the following conditions holds: (1) the decision model returns a stop signal $\sigma_t = 1$; (2) the current node has no child nodes; or (3) the search reaches the maximum depth $D_{\max}$. The resulting path is

\begin{equation}
p(x) = (v_0, v_1, \ldots, v_L), \quad v_0 = v_{\mathrm{root}}.
\end{equation}

\subsubsection{Path-Level Evidence Aggregation}

After a path is obtained, the system does not rely only on the evidence attached to the terminal node. In regulatory classification, the final decision is often supported by conditions introduced at multiple hierarchy levels. For example, a parent node may define the overall scope of a category, an intermediate node may contribute an exclusion or note, and the terminal node may provide the most specific class definition. We therefore aggregate evidence from all nodes along the selected path:

\begin{equation}
\mathcal{S}(x)
= \operatorname{MergeTop}_{v \in p(x)}
\mathcal{S}(v \mid x).
\end{equation}

The $\operatorname{MergeTop}$ operation turns these node-level evidence sets into a compact path-level evidence pool. It removes duplicates, ranks snippets by relevance, and keeps the most useful evidence for the current decision. The resulting set $\mathcal{S}(x)$ supports final decision making, rationale generation, and confidence assessment.

The final decision generator $G_\theta$ is used after the search path has already been determined. It does not select another child node or change the hierarchy path. Instead, it takes the original input $x$, the selected path $p(x)$, and the aggregated evidence $\mathcal{S}(x)$, and converts them into a structured final output. Reaching a terminal node is not always enough: if the evidence is insufficient, conflicting, or weak, $G_\theta$ can flag the case as low-confidence instead of over-committing:

\begin{equation}
\hat{y} = G_\theta(x, p(x), \mathcal{S}(x)),
\end{equation}
where $\hat{y}$ instantiates the final output $y(x)$ with the predicted class, confidence, supporting evidence, and a rationale. In this sense, $F_\theta$ is responsible for local branch selection during search, whereas $G_\theta$ is responsible for final support verification and output formatting after the path-level evidence has been aggregated.

\subsection{Optional Fast Anchor Initialization}

The local hierarchical search above follows a strict top-down procedure. This preserves the regulatory hierarchy, but it may require several steps before reaching a relevant subtree. As an optional acceleration module, we can first perform a global retrieval over all indexed nodes:

\begin{equation}
g_{v \mid x}
= \cos\big(\phi(q^{\mathrm{ret}}), \mathbf{h}_v\big),
\quad v \in \mathcal{V}.
\end{equation}

The top-$M$ globally retrieved nodes are retained as the initial candidate set:

\begin{equation}
    \mathcal{C}^{\mathrm{glob}}(x)
= \operatorname{TopM}_{v \in \mathcal{V}}
g_{v \mid x}.
\label{eq:25}
\end{equation}

Here, $\operatorname{TopM}$ uses the same candidate-size parameter $topM$ as the local retrieval operator, but applies it to the full node set $\mathcal{V}$ rather than only to the children of the current node. Thus, $\mathcal{C}^{\mathrm{glob}}(x)$ plays the same role as the local candidate set $\mathcal{C}_t(x)$, except that it is obtained from global retrieval. The retrieved candidates are then passed to the same candidate package construction step, where each candidate node is enriched with structured regulatory fields and retrieved evidence before entering the next-hop decision process. After a global candidate is selected as the starting anchor, the system continues local hierarchical search from that node. If the full path is needed, the ancestor chain from the root to the selected anchor is prepended to the locally refined path, so the result remains consistent with the hierarchy.

\section{Experiments and Results}

This section presents the benchmark construction, baseline methods, and experimental settings used to evaluate the proposed constraint-aware hierarchical search framework.

\subsection{Dataset Construction}

To evaluate regulation-driven fine-grained hierarchical classification, we construct four benchmark datasets from three types of official or standard regulatory sources: the Chinese customs HS commodity and heading notes, the dual-use item export control list, and the classification standard for basic education equipment. Table \ref{tab:dataset} summarizes these datasets and their regulatory hierarchies. These sources are suitable for our task because they all define hierarchical category systems and require classification decisions to follow explicit descriptions, technical conditions, exclusions, and boundary rules.

HS-Simple and HS-Hard are derived from the same HS regulatory source, but they are separated by instance difficulty. HS-Simple contains relatively direct commodity classification cases. In these cases, the correct code can usually be identified from category descriptions and common product attributes. HS-Hard focuses on boundary-sensitive cases. The instances are surface-similar to neighboring categories but belong to different HS codes, and the correct decision depends on fine-grained attributes such as composition, usage, form, size, thickness, or threshold conditions.

To make the difficulty more concrete, Table \ref{tab:dataset_examples} shows four representative HS examples used during benchmark construction. These examples are intentionally surface-similar: they share nearly identical product types, forms, and manufacturing descriptions. However, their final labels differ because of fine-grained regulatory boundaries such as alloy composition or thickness thresholds. This makes the benchmark challenging for methods that mainly rely on semantic similarity.

\begin{table*}[t]
\centering
\caption{Summary of the four constructed datasets. HS-Simple and HS-Hard are both derived from the HS tariff corpus, but HS-Hard contains more neighboring-code cases with subtle attribute differences.}
\begin{tabular}{l p{4.2cm} l l p{5.2cm}}
\toprule
\textbf{Dataset} & \textbf{Source Regulation or Standard} & \textbf{Difficulty} & \textbf{\# Instances} & \textbf{Regulatory Hierarchy} \\
\midrule
\textbf{HS-Simple}
& Chinese customs HS commodity and heading notes
& 2 & 370
& Four-level HS structure: section, chapter, heading, and subheading \\
\textbf{HS-Hard}
& Chinese customs HS commodity and heading notes
& 5 & 370
& Four-level HS structure: section, chapter, heading, and subheading \\
\textbf{Dual-Use List}
& Dual-use item export control list
& 4 & 270
& Four-level code structure: industry category, item type, controlled item, and technical subitem \\
\textbf{Educational Equipment}
& Classification and code standard for basic education equipment
& 3 & 300
& Five-segment, three-level code: major category, middle category, minor category, sequence number, and model \\
\bottomrule
\end{tabular}
\label{tab:dataset}
\end{table*}

\begin{table*}[t]
\centering
\caption{Representative HS examples with boundary-sensitive differences. The descriptions are semantically similar, but the correct codes differ because of local thresholds or composition-based rules.}
\label{tab:dataset_examples}
\small
\setlength{\tabcolsep}{5pt}
\begin{tabularx}{\textwidth}{l X c}
\toprule
Example & Surface-Similar Description & Gold Code \\
\midrule
Aluminum strip A
& Rectangular non-alloy aluminum strip for electronic parts; thickness 0.4\,mm, width 200\,mm; aluminum content 99.5672--99.589\%.
& 760611 \\

Aluminum strip B
& Aluminum coil/strip for can-body material; thickness 0.245\,mm, width 1766.1\,mm; alloy grade 3104 with Mn/Mg/Cu/Si/Fe additives.
& 760612 \\

Stainless steel plate A
& Hot-rolled stainless steel plate, 1D, not pickled; size 12.0 $\times$ 2000 $\times$ 7100\,mm.
& 721921 \\

Stainless steel plate B
& Hot-rolled stainless steel plate, 1D, not pickled; size 6.0 $\times$ 2100 $\times$ 9000\,mm.
& 721922 \\
\bottomrule
\end{tabularx}
\end{table*}

For each dataset, input instances are written as natural language classification requests. Each instance describes an object, product, item, or piece of equipment using attributes such as material, function, usage, composition, technical parameters, model, and other domain-specific conditions. The gold annotation contains the final regulatory code and the corresponding hierarchy path.

To ensure annotation reliability, benchmark construction follows an expert-in-the-loop validation process. We first construct natural language instances and assign preliminary labels according to the source regulatory documents. For each instance, the gold label is determined by following the official regulatory hierarchy from coarse to fine. Annotators first identify the applicable upper-level category and then resolve lower-level branches using the source descriptions, notes, exclusions, definitions, and threshold conditions. The sequence of selected nodes forms the gold path, and its terminal node is used as the gold label. Domain experts from an industry partner with practical experience in regulatory classification and compliance services then validate both the path and the label. An instance is accepted only when every path transition is reachable in the regulatory tree and the decisive conditions for the terminal node are supported by the source text. Instances with ambiguous descriptions, missing decisive attributes, insufficient evidence, or disputed path assignments are revised or removed until a consistent annotation is obtained.

The constructed benchmarks are designed to test whether a model can move beyond surface-level semantic matching. A successful system must identify the relevant regulatory branch, compare fine-grained conditions, reject legally invalid neighboring classes, and provide decisions grounded in the source documents.

\subsection{Baseline Methods}

We compare our method with representative LLM-based and retrieval-augmented baselines. These methods cover direct generation, conventional chunk-based RAG, graph-enhanced RAG, and recent RAG variants designed to improve retrieval efficiency, graph-based retrieval, or multi-step evidence use.

We focus on LLM-based baselines because regulation-driven classification is not a standard supervised classification setting with a fixed set of short labels and abundant labeled training data. Traditional classifiers would require task-specific training data for each regulatory domain and usually cannot directly use newly updated legal texts or generate evidence-grounded rationales. In contrast, LLMs and RAG systems provide a stronger and more realistic comparison, since they can operate over natural language regulations in a zero-shot or retrieval-augmented manner.

\textbf{LLM direct answer.} This baseline directly prompts an LLM to predict the final class from the input description and task instruction, without retrieving external regulatory evidence. It tests whether the model can solve regulation-driven classification using only its parametric knowledge and general reasoning ability.

\textbf{Flat Evidence RAG + Candidate Label Rerank.} Following the standard retrieval-augmented generation paradigm \cite{lewis2020rag,gao2023rag}, this baseline splits regulatory documents into evidence chunks and retrieves the top relevant chunks from the full regulatory corpus for each query. Each retrieved chunk is linked to its associated regulatory node or final code, and the linked codes are collected as candidate labels. The LLM then receives the input description, retrieved evidence chunks, and candidate labels, and reranks the candidates to select the final class. This baseline controls for the benefit of evidence retrieval and explicit candidate labels, while keeping the retrieval and reranking process flat: it does not restrict candidates to the children of the current node or enforce stepwise parent-child decisions.

\textbf{LightRAG (hybrid).} LightRAG builds lightweight graph-based indexes and combines local and global retrieval signals to improve retrieval-augmented generation efficiency \cite{guo2024lightrag}. We use its hybrid retrieval mode as a baseline to test whether a general-purpose efficient RAG framework can handle fine-grained regulatory classification.

\textbf{GraphRAG.} GraphRAG constructs graph-structured representations from the corpus and performs query-focused retrieval over graph communities and textual evidence \cite{edge2024graphrag}. This baseline evaluates whether graph-based knowledge organization can improve classification decisions that depend on long regulatory documents and cross-document relations.

\textbf{LinearRAG.} LinearRAG is a graph retrieval-augmented generation method designed to perform linear graph retrieval on large-scale corpora \cite{zhuang2025linearrag}. We include it to compare our hierarchical regulatory search with a recent graph-based RAG approach that emphasizes scalable retrieval over structured knowledge.

\textbf{MA-RAG.} MA-RAG introduces multi-agent retrieval-augmented generation with collaborative chain-of-thought reasoning \cite{nguyen2025marag}. This baseline is used to examine whether multi-agent reasoning over retrieved evidence can improve fine-grained regulatory classification, especially for cases requiring multi-step comparison among competing classes.

\begin{table*}[t]
\centering
\caption{Final-label accuracy of all baseline methods and our method on the four benchmark datasets. Values are mean accuracy over three runs, with sample standard deviations shown in parentheses.}
\small
\setlength{\tabcolsep}{3pt}
\begin{tabularx}{\textwidth}{>{\raggedright\arraybackslash}p{0.30\textwidth}*{4}{>{\centering\arraybackslash}X}}
\toprule
\textbf{Method} & \textbf{HS-Hard} & \textbf{HS-Simple} & \textbf{Dual-Use List} & \textbf{Educational Equipment} \\
\midrule
LLM direct answer     & 17.93 (0.56)\% & 36.29 (2.10)\% & 3.21 (0.93)\%  & 0.56 (0.20)\%  \\
LightRAG (hybrid)     & 29.82 (2.86)\% & 70.31 (4.77)\% & 74.07 (1.86)\% & 82.22 (0.84)\% \\
Flat Evidence RAG + Candidate Label Rerank & 31.44 (1.80)\% & 72.76 (0.95)\% & 78.04 (1.40)\% & 85.55 (0.39)\% \\
GraphRAG              & 21.14 (7.96)\% & 38.26 (2.43)\% & 10.49 (1.13)\% & 81.78 (1.83)\% \\
LinearRAG             & 40.72 (10.48)\% & 80.47 (1.83)\% & 54.69 (2.63)\% & 21.22 (6.06)\% \\
MA-RAG                & 47.93 (1.74)\% & 66.22 (1.58)\% & 20.99 (1.07)\% & 55.56 (3.47)\% \\
Ours                  & \textbf{58.29 (0.87)\%} & \textbf{90.44 (1.36)\%} & \textbf{78.64 (0.93)\%} & \textbf{96.11 (1.17)\%} \\
\bottomrule
\end{tabularx}
\label{tab:main_results}
\end{table*}

\subsection{Experimental Setup}

\subsubsection{Evaluation Metrics}

We evaluate performance from two complementary perspectives: final-label accuracy, which measures whether the predicted fine-grained regulatory class is correct, and level-wise accuracy, which measures whether the hierarchical decision path remains correct at each decision level.

We use final-label accuracy as the primary evaluation metric. It measures whether the predicted fine-grained regulatory class exactly matches the gold label for each input instance. For a dataset with $N$ test instances, the metric is defined as

\begin{equation}
\mathrm{Acc}
= \frac{1}{N}\sum_{i=1}^{N}\mathbf{1}(\hat{y}_i = y_i),
\end{equation}
where $\hat{y}_i$ is the predicted final label and $y_i$ is the gold final label.

Final-label accuracy alone does not reveal at which hierarchy level an error first occurs. We therefore also report level-wise accuracy, which evaluates the intermediate hierarchy-constrained decision process. For an instance $x_i$, let $y_i^{(k)}$ and $\hat{y}_i^{(k)}$ denote the gold node and predicted node at hierarchy level $k$, respectively. A prediction is counted as correct at level $k$ if $\hat{y}_i^{(k)} = y_i^{(k)}$. The level-wise accuracy at level $k$ is defined as

\begin{equation}
\mathrm{Acc}^{(k)}
= \frac{1}{N}\sum_{i=1}^{N}\mathbf{1}\big(\hat{y}_i^{(k)} = y_i^{(k)}\big),
\end{equation}
where $N$ is the number of test instances. For the HS datasets, we report the first three decision levels, corresponding to the coarse-to-fine progression from upper-level routing to deeper branch discrimination.

The final decision generator is designed to flag low-confidence cases when the aggregated regulatory evidence is insufficient. However, all benchmark instances in this study are constructed with gold regulatory labels and are therefore treated as answerable cases. We consequently evaluate classification performance using final-label and level-wise accuracy, and leave systematic evaluation of abstention or unresolved outputs to future work.

In addition to the automatic accuracy score, we manually inspect a subset of predictions to analyze common error types, including wrong branch selection, missed exclusion clauses, incorrect threshold comparison, and confusion between neighboring fine-grained classes. This analysis is used only to interpret model behavior and is not reported as a separate automatic evaluation metric.

\subsubsection{Implementation Details}

For retrieval-based methods, we use the same indexed regulatory corpus and the same embedding backend whenever possible, so that differences mainly reflect the decision strategy rather than the retrieval backend. All text embeddings are produced by Alibaba's \textit{text-embedding-v4}\footnote{https://aimlapi.com/models/qwen-text-embedding-v4} model, and the embedding dimensionality is fixed to 1024 for all datasets and methods. The number of retrieved candidates is selected on the development set. For hierarchical methods, we use the same maximum search depth and stopping criteria unless otherwise specified. The Flat Evidence RAG + Candidate Label Rerank baseline uses the same evidence chunks and label associations as the indexed corpus, but performs one-shot global evidence retrieval and global candidate reranking instead of hierarchy-scoped local branch selection.

For LLM-based components, all methods use DeepSeek V3.2 \cite{liu2025deepseek} as the answer generation and decision model. We use a fixed prompt template for each method and keep decoding settings consistent across datasets. All methods are evaluated on the same test split in three independent runs. We report the mean accuracy and sample standard deviation across the three runs for each domain. Additional implementation details of our method and the baseline systems are provided in Appendix~\ref{app:ours} and Appendix~\ref{app:baselines}, respectively.

\subsection{Main Results}

Table \ref{tab:main_results} reports final-label accuracy on the four benchmark datasets, using the mean and standard deviation over three runs. Direct LLM prediction performs poorly on most datasets, especially on the Dual-Use List and Educational Equipment datasets. This indicates that regulation-driven classification cannot be reliably solved by parametric knowledge alone, since the final label often depends on domain-specific regulatory descriptions and fine-grained boundary conditions.

Among the baselines, MA-RAG obtains the strongest result on HS-Hard, with 47.93 $\pm$ 1.74\%, while LinearRAG performs best on HS-Simple, with 80.47 $\pm$ 1.83\%. Flat Evidence RAG + Candidate Label Rerank is the strongest baseline on Dual-Use List and Educational Equipment, reaching 78.04 $\pm$ 1.40\% and 85.55 $\pm$ 0.39\%, respectively. This pattern suggests that global evidence retrieval and candidate reranking can work well when the retrieved evidence is distinctive, but they are less reliable on HS-Hard, where the model must distinguish among neighboring headings or subheadings under local regulatory conditions.

Our method achieves the best mean performance on all four datasets, with 58.29 $\pm$ 0.87\% on HS-Hard, 90.44 $\pm$ 1.36\% on HS-Simple, 78.64 $\pm$ 0.93\% on Dual-Use List, and 96.11 $\pm$ 1.17\% on Educational Equipment. The gains over the strongest baseline are substantial on HS-Hard, HS-Simple, and Educational Equipment, reaching 10.36, 9.97, and 10.56 percentage points, respectively. On Dual-Use List, the gain is smaller at 0.60 percentage points, indicating that flat evidence retrieval already covers many cases when the candidate evidence is distinctive. Overall, the results suggest that hierarchy-scoped candidate comparison and structured regulatory fields are most beneficial when the label space contains fine-grained neighboring categories and rule-based boundary conditions.

\subsection{Level-Wise Accuracy Analysis}

Level-wise accuracy complements final-label accuracy by revealing at which hierarchy level the error first occurs. In HS classification, this distinction is important because a wrong prediction may either result from an early routing mistake at the section or chapter level, or from a later confusion among fine-grained neighboring headings and subheadings. We therefore report level-wise accuracy on the two HS datasets to analyze whether the model preserves correct decisions as the search proceeds deeper into the hierarchy. Table \ref{tab:levelwise_acc} reports the results of three runs together with their average and standard deviation.

\begin{table*}[t]
\centering
\small
\setlength{\tabcolsep}{2.5pt}
\caption{Level-wise accuracy of our method on the two HS datasets. We report three runs, the average, and the sample standard deviation for each hierarchy level; standard deviations are shown in parentheses.}
\begin{tabular}{lcccccccccc}
\toprule
\textbf{Level} & \multicolumn{5}{c}{\textbf{HS-Simple}} & \multicolumn{5}{c}{\textbf{HS-Hard}} \\
\cmidrule(lr){2-6} \cmidrule(lr){7-11}
& \textbf{Run 1} & \textbf{Run 2} & \textbf{Run 3} & \textbf{Avg.} & \textbf{(Std.)} & \textbf{Run 1} & \textbf{Run 2} & \textbf{Run 3} & \textbf{Avg.} & \textbf{(Std.)} \\
\midrule
Level 1 & 97.58\% & 97.31\% & 95.97\% & \textbf{96.95\%} & (0.86)\% & 83.24\% & 84.59\% & 86.49\% & \textbf{84.77\%} & (1.63)\% \\
Level 2 & 96.77\% & 96.41\% & 92.74\% & \textbf{95.31\%} & (2.23)\% & 71.89\% & 72.70\% & 74.32\% & \textbf{72.97\%} & (1.24)\% \\
Level 3 & 91.67\% & 90.66\% & 88.98\% & \textbf{90.44\%} & (1.36)\% & 57.30\% & 58.92\% & 58.65\% & \textbf{58.29\%} & (0.87)\% \\
\bottomrule
\end{tabular}
\label{tab:levelwise_acc}
\end{table*}

Several observations can be drawn from Table \ref{tab:levelwise_acc}. On HS-Simple, the model maintains high accuracy across all three levels, with average scores of 96.95\%, 95.31\%, and 90.44\%, respectively. This indicates that the proposed constraint-aware search framework reliably identifies the upper-level route and remains effective when the decision moves to deeper branches.

On HS-Hard, the level-wise accuracy decreases more sharply, from 84.77\% at Level 1 to 72.97\% at Level 2 and 58.29\% at Level 3. This trend suggests that difficult HS cases not only make upper-level routing harder, but also amplify error accumulation as the search descends the hierarchy. The main remaining challenge lies in distinguishing closely related neighboring headings or subheadings whose surface descriptions are similar but whose regulatory conditions differ in thresholds, exclusions, or applicability clauses.

Overall, the level-wise analysis complements the final-label results by showing where errors occur along the path. The method captures the high-level taxonomy well, while the remaining errors are concentrated at deeper fine-grained decision stages, especially on HS-Hard. This finding is consistent with the motivation of regulation-driven hierarchical classification: correct high-level localization is necessary, but robust performance ultimately depends on resolving subtle local rule boundaries among valid sibling candidates.

\subsection{Ablation Study}

\subsubsection{Effect of Fast Anchor Initialization}

We further conduct an ablation study to evaluate the effect of the optional fast anchor initialization module. The experiment is performed on the HS-Hard dataset, where the regulatory tree has a depth of 4. We set the candidate size to $topM=8$ (Equation \ref{eq:25}) for both settings. The variant without fast anchor initialization starts the search from the root node and follows the standard top-down hierarchical search procedure. The variant with fast anchor initialization first retrieves a global candidate set from the full node index, passes these candidates through the same candidate package construction and next-hop decision process, and then continues local hierarchical search from the selected anchor.

\begin{figure}
    \centering
    \includegraphics[width=\linewidth]{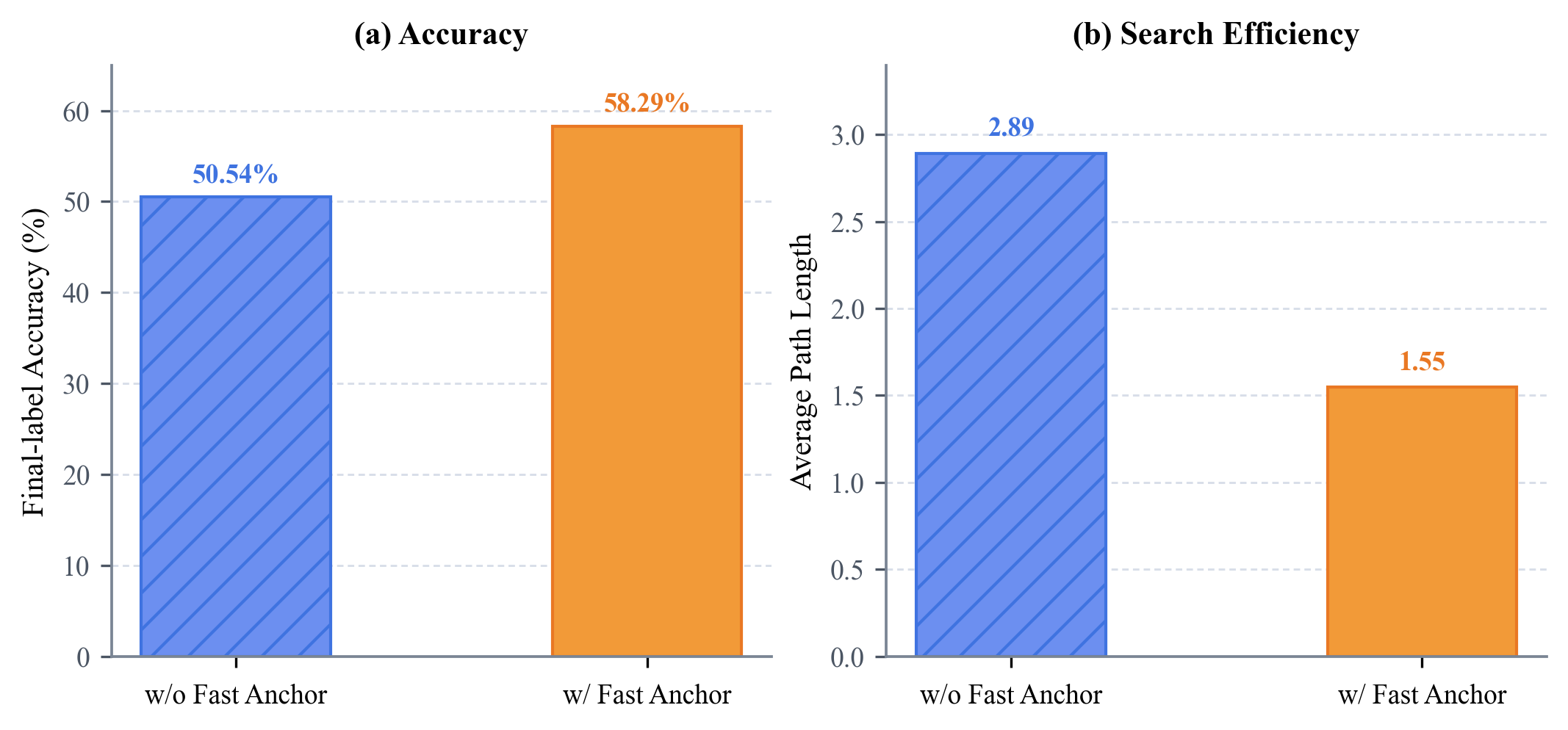}
    \caption{Ablation study on HS-Hard comparing Fast Anchor initialization with a baseline without it. Fast Anchor improves final-label accuracy from 50.54\% to 58.29\% and reduces the average search path length from 2.89 to 1.55, indicating both higher classification performance and more efficient search.}
    \label{fig:ab1}
\end{figure}

Figure \ref{fig:ab1} shows that fast anchor initialization improves final-label accuracy from 50.54\% to 58.29\%, while reducing the average path length from 2.89 to 1.55. This indicates that the anchor module helps the model enter the relevant regulatory region earlier and reduces unnecessary upper-level traversal. The shorter path also improves accuracy, suggesting that avoiding long top-down searches can reduce error accumulation.

\subsubsection{Effect of Candidate Set Size $topM$}

We further study the impact of the candidate set size $topM$ (Equation \ref{eq:25}), which controls both the number of globally retrieved candidates for fast anchor initialization and the number of locally retrieved child candidates during hierarchical search. We evaluate $topM \in \{3, 5, 8\}$ on HS-Simple and HS-Hard.

\begin{figure}
    \centering
    \includegraphics[width=\linewidth]{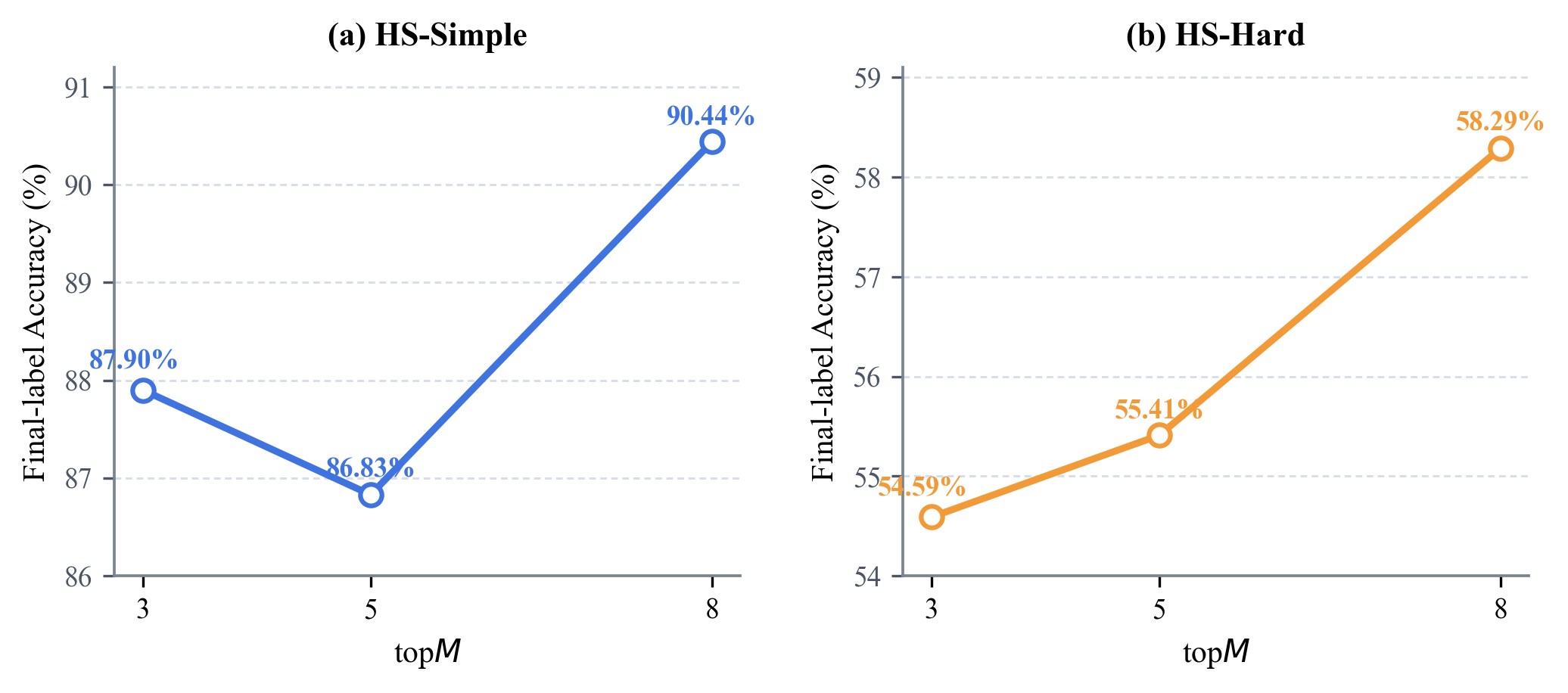}
    \caption{Sensitivity of candidate set size $topM$ on HS benchmarks. HS-Simple shows limited sensitivity to $topM$ (87.90\%, 86.83\%, and 90.44\% for $topM$ = 3, 5, and 8), while HS-Hard benefits from larger $topM$ (54.59\% $\rightarrow$ 58.29\%), suggesting that harder instances require higher candidate coverage to keep the correct next hop among candidates.}
    \label{fig:ab2}
\end{figure}

As shown in Figure \ref{fig:ab2}, performance on HS-Simple remains relatively stable across different $topM$ settings, with a maximum gap of 3.61 percentage points. This suggests that the decisive evidence for easier cases is concentrated and that a small candidate set is often sufficient to cover the correct branch.

On HS-Hard, accuracy improves consistently as $topM$ increases (58.29\% at $topM=8$ vs. 54.59\% at $topM=3$). This indicates that harder instances benefit from higher candidate coverage: with fine-grained neighboring headings/subheadings, a small $topM$ can miss the correct child early, causing the search to commit to an incorrect branch. Retaining more candidates increases the probability that the correct next hop remains available for subsequent evidence-based comparison.

Overall, we set $topM=8$ as the default candidate set size in our HS experiments to balance retrieval coverage with computational cost.

\subsubsection{Candidate Field Ablation in Branch Selection}

We ablate the candidate node fields exposed to the LLM during the branch selection stage (the next-hop decision $F_{\theta}$). This experiment evaluates how progressively enriching the candidate package $\Gamma(u)$ affects path selection and final-label accuracy. It also studies the contribution of explicit regulatory fields, including notes, exclusion clauses, and definitions: these fields are removed in the first three variants and are added only in the Full Candidate setting. Unless otherwise specified, we use $topM=8$ for local candidate retrieval.

We consider four settings with increasing information:

\textbf{Title-only.} This is the weakest-information baseline, testing whether the model can match the query to the correct child using only a short semantic label and structural type.

\textbf{Title + Evidence.} Adds evidence excerpts (the retrieved snippet set $\mathcal{S}(u \mid x)$) to validate whether concise, task-relevant evidence helps distinguish neighboring candidates.

\textbf{Title + Text.} Adds the node body text to test whether the full regulatory text improves understanding of table-like blocks and provides useful context beyond short evidence.

\textbf{Full Candidate.} Uses \texttt{code + title + text + notes + exclusions + definitions + evidence}. Compared with the previous variants, this setting explicitly exposes the regulatory fields that often determine applicability, such as local notes, exclusion clauses, and definitional provisions.

The results show that adding evidence excerpts yields a large gain over Title-only, especially on HS-Simple, supporting our design choice that concise, retrieval-grounded evidence provides actionable signals for local comparisons among siblings. Adding the full node text provides mixed benefit: it can supply extra context, but may also introduce irrelevant details and make it harder to focus on decisive conditions. The Full Candidate setting achieves the best accuracy on both HS-Simple and HS-Hard. Compared with Title + Text, it improves accuracy by 9.53 and 2.34 percentage points, respectively; compared with Title + Evidence, it improves accuracy by 4.15 and 2.34 percentage points. These gains indicate that explicit regulatory fields, especially notes, exclusions, and definitions, provide useful constraints beyond title matching, evidence retrieval, or raw node text.

\begin{figure}
    \centering
    \includegraphics[width=\linewidth]{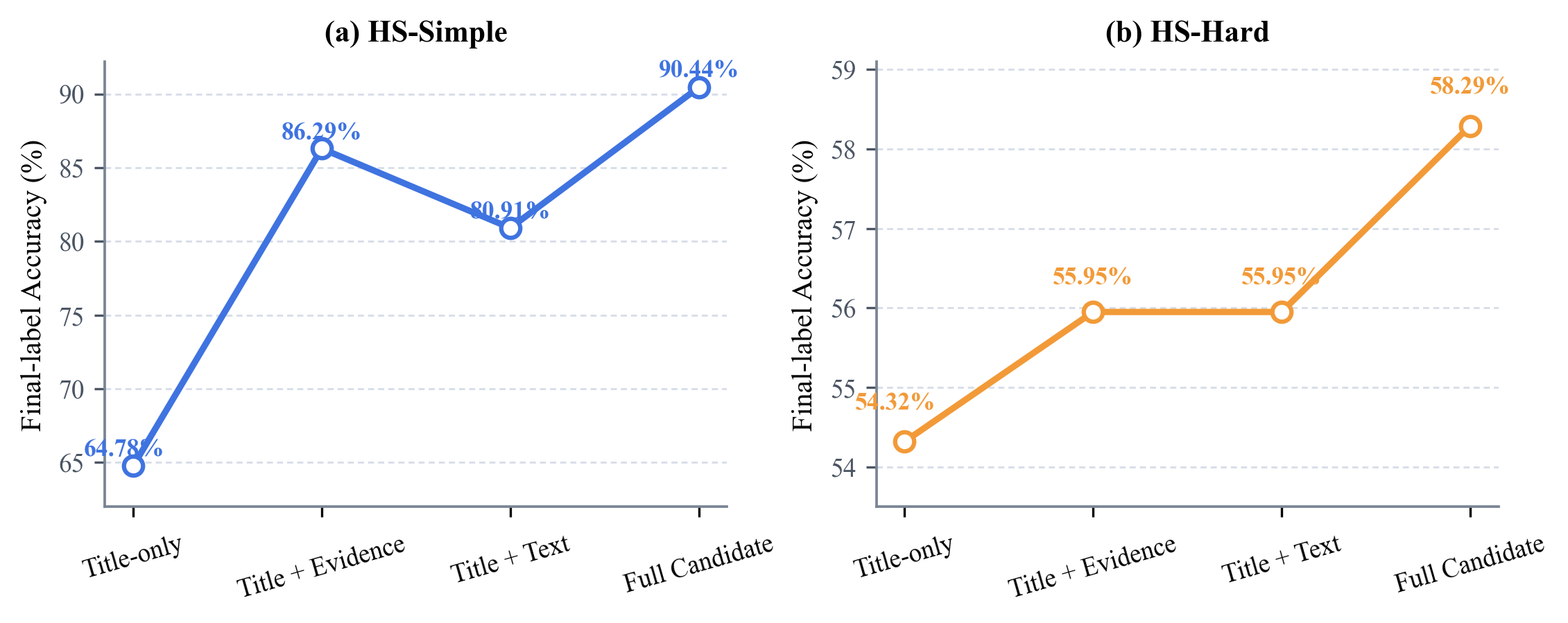}
    \caption{Candidate field ablation for next-hop branch selection ($F_{\theta}$) with $topM=8$. We progressively enrich the candidate package $\Gamma(u)$ from \textbf{Title-only} to \textbf{Full Candidate} and report final-label accuracy on HS-Hard and HS-Simple. The first three variants omit explicit regulatory fields such as notes, exclusions, and definitions, while the Full Candidate variant includes them.}
    \label{fig:ab3}
\end{figure}

\subsection{Case Study}

The aggregate results across multiple regulatory classification scenarios show that our method improves final-label accuracy, but accuracy alone does not reveal why different methods succeed or fail. We therefore conduct a qualitative case study to examine whether a method can identify the decisive local rule, reject semantically plausible but invalid neighboring classes, and ground the final decision in applicable evidence. We choose two representative examples from HS-Hard because this dataset contains many fine-grained neighboring codes and explicit subheading-level conditions, making it suitable for illustrating boundary-sensitive errors. Both cases are difficult because the surface description matches several semantically related HS codes, while the correct decision depends on applying a specific subheading condition or threshold.

\subsubsection{Case 1: LLDPE Copolymer}

\textbf{Key features.} The product is a linear low-density polyethylene (LLDPE) resin in primary granule form. It is mainly composed of an ethylene and 1-octene copolymer, with ethylene accounting for 85\%, 1-octene accounting for 15\%, and a density of 0.907.

\textbf{Gold code.} The correct 6-digit HS code is 390140.

\textbf{Baseline errors.} The direct LLM and GraphRAG predict 390110, mainly because they over-rely on the surface name LLDPE-polyethylene and the low-density signal. LightRAG predicts 390190, indicating that it captures the copolymer issue but fails to select the more specific ethylene-$\alpha$-olefin subheading.

\textbf{Decisive rule.} The product should not be classified merely by the surface phrase "polyethylene". The decisive condition is the subheading description under heading 39.01: ethylene-$\alpha$-olefin copolymers with a density below 0.94.

\textbf{Analysis.} Our method retrieves evidence indicating that heading 39.01 covers LLDPE and then compares the candidate subheadings under the local hierarchy. It identifies 1-octene as an $\alpha$-olefin, verifies that the product is an ethylene-$\alpha$-olefin copolymer, and checks that the density 0.907 satisfies the below-0.94 condition. Therefore, the method selects 390140. This case shows that semantically relevant evidence is not enough: the system must apply the correct subheading-level condition within the local regulatory hierarchy.

\subsubsection{Case 2: Self-Adhesive Reflective Film}

\textbf{Key features.} The product is a capsule-type reflective film that is self-adhesive, in rolls, and wider than 20 cm.

\textbf{Baseline errors.} LightRAG predicts 392010, suggesting that it follows the general semantic cue of plastic film but misses the more specific self-adhesive heading 39.19. GraphRAG drifts toward 392190 or other general plastic sheet/film categories, again failing to apply the width threshold under the correct heading. Although the direct LLM reaches a related 391990-level answer, it outputs an overly specific 8-digit code and relies more on general product knowledge than on explicit retrieved evidence.

\textbf{Decisive rule.} The decisive rule is the width condition under heading 39.19, which covers self-adhesive plastic plates, sheets, film, foil, tape, strip, and other flat shapes, whether or not in rolls. Within this heading, code 391910 applies only to rolls with a width not exceeding 20 cm, while 391990 covers other cases.

\textbf{Analysis.} Our method retrieves this local rule and uses the key attributes "self-adhesive", "in rolls", and "width > 20 cm" to reject 391910 and select 391990. It also avoids being distracted by non-decisive attributes such as "reflective", "capsule-type", and "UL". This case shows the importance of constraining the comparison to valid local candidates and using rule-grounded evidence for next-hop decisions.

Overall, these cases illustrate that failures in HS-Hard are often not caused by an absence of relevant text, but by the inability to determine which local rule is decisive. Baseline methods tend to follow surface semantic cues, retrieve broadly related plastic categories, or miss subheading-level thresholds. Our constraint-aware hierarchical search keeps the decision within the current regulatory scope, attaches evidence to candidate nodes, and enables the model to compare legally meaningful conditions before selecting the final code.

\section{Conclusion}

In this paper, we introduced \textbf{Regulation-driven Fine-grained Hierarchical Classification}, a task that requires assigning an external instance to a fine-grained class defined by a rule-governed hierarchy. Unlike conventional text classification or standard RAG settings, the correct decision must satisfy hierarchical validity, explicit regulatory constraints, and boundary-sensitive conditions. To study this task, we constructed four benchmark datasets from representative regulation-intensive classification scenarios with expert-in-the-loop validation.

We proposed a constraint-aware hierarchical search framework that converts regulatory documents into a searchable tree, retrieves valid local candidate nodes at each decision step, and uses structured candidate information with evidence snippets to guide next-hop selection. The resulting prediction includes not only the final class, but also a hierarchy-consistent path and supporting evidence. Experiments over three runs show that our method outperforms or matches direct LLM prediction and retrieval-augmented baselines across datasets, with the largest gains on HS-Hard, HS-Simple, and Educational Equipment. Further ablations demonstrate the importance of fast anchor initialization, candidate set size, and candidate field design, while case studies illustrate how the proposed method handles rule-based boundary conditions that mislead general LLM and RAG methods.

Overall, our results suggest that regulation-driven classification should be treated as a constraint-aware hierarchical decision problem rather than flat semantic matching. Future work will extend the evaluation to broader regulatory domains, more comprehensive repeated trials, and stronger robustness analysis, while further improving the scalability of hierarchical search in large regulatory trees. We hope this task formulation, benchmark suite, and search framework can support future research on reliable, auditable, and rule-grounded classification systems for high-stakes regulatory domains.

\section*{Data and Code Availability}

The source code and benchmark datasets used in this study are available at \url{https://github.com/tanwei20011111/Constraint-Aware-Hierarchical-Search-#tree-construction-h}. The dataset files are provided under the \url{https://github.com/tanwei20011111/Constraint-Aware-Hierarchical-Search-/tree/main/data} directory.

\appendix
\section{Implementation Details of the Proposed Method}
\label{app:ours}

This appendix provides additional implementation details for the proposed constraint-aware hierarchical search framework. The implementation has two stages: offline regulatory tree construction and online hierarchy-constrained search. The offline stage produces the regulatory tree, node index, evidence index, and node-evidence associations used by all experiments. The online stage takes a natural language instance as input and returns the predicted code, hierarchy path, supporting evidence, confidence, and rationale. All components use the same LLM and embedding backend described in the experimental setup.

\textbf{Offline tree construction.} The tree builder follows a read--extract--materialize--attach workflow. First, the system reads the source text and document-level metadata, such as title, source file, and page range. The metadata is not used to infer the hierarchy. It is retained so that each node and evidence snippet can be traced back to the original regulatory source.

Second, the document text is split into structure-aware text blocks. These blocks are different from evidence chunks. A text block is the input given to the LLM for hierarchy extraction. It should contain enough local context for the model to recognize headings, subheadings, notes, and explanatory provisions. An evidence chunk is a short source span used later for retrieval and citation. Evidence chunks support auditability, but they do not decide parent-child relations.

Third, the LLM reads each text block and returns structured regulatory units. Each unit contains fields such as code, title, node type, original text, notes, exclusions, definitions, and \texttt{children}. The \texttt{children} field is the key output for tree construction. It states which lower-level units belong under the current unit. For example, a simplified extraction from HS Chapter 18 can be represented as follows:

\begin{verbatim}
Chapter 18
  -> 18.01, 18.02, 18.03, 18.06
18.03
  -> 1803.10, 1803.20
18.06
  -> 1806.10, 1806.20, 1806.31, 1806.32, 1806.90
1806.20
  -> Note for subheading 1806.20
1806.31
  -> Note for subheading 1806.31
\end{verbatim}

The returned object can be understood as a nested structure:

\begin{verbatim}
{
  "code": "18",
  "title": "Chapter 18 Cocoa and cocoa preparations",
  "children": [
    {
      "code": "18.06",
      "title": "Chocolate and other food preparations...",
      "children": [
        {
          "code": "1806.20",
          "title": "Other preparations in blocks...",
          "children": [{"title": "Note for subheading 1806.20"}]
        },
        {
          "code": "1806.31",
          "title": "Filled",
          "children": [{"title": "Note for subheading 1806.31"}]
        }
      ]
    }
  ]
}
\end{verbatim}

Fourth, the implementation materializes this nested \texttt{children} structure into the global tree. It first creates a root and document-level node, denoted here as \texttt{doc::input}, and then recursively writes each extracted unit into the tree. In simplified form, the procedure is

\begin{verbatim}
materialize(parent, units):
    for unit in units:
        node = create_node(unit)
        parent.children.append(node)
        materialize(node, unit.children)
\end{verbatim}

Thus, once a node is attached to its parent, the node immediately becomes the parent for the next level. The same operation is repeated from coarse to fine until all nested children have been written. For the 18.06 branch above, the materialized tree is

\begin{verbatim}
doc::input
  -> Chapter 18 Cocoa and cocoa preparations
       -> 18.06 Chocolate and other food preparations...
            -> 1806.20 Other preparations in blocks...
                 -> Note for subheading 1806.20
            -> 1806.31 Filled
                 -> Note for subheading 1806.31
\end{verbatim}

Equivalently, recursive materialization creates edges such as $(\text{Chapter 18}, 18.06)$, $(18.06, 1806.20)$, and $(18.06, 1806.31)$. The implementation does not re-infer these parent-child relations from evidence chunks. It only normalizes, deduplicates, validates, and materializes the hierarchy already expressed by \texttt{children}. Dataset-specific extraction rules, such as hierarchy depth and code format, are generated from representative source samples and injected into the tree-construction prompt before extraction.

Fifth, after the nodes have been created, evidence chunks are associated with the corresponding nodes. For example, a chunk containing the wording of heading 18.06 is attached to the 18.06 node, while a chunk explaining the meaning of ``filled'' is attached to the note under 1806.31. When a fine-grained node does not have enough direct evidence, inherited evidence from its ancestors is retained as supplementary context. The offline stage finally saves the tree structure, including node identifiers and \texttt{children} relations, and the vector representations used by the node and evidence indexes.

\textbf{Embedding and indexing.} We build two vector indexes after tree construction. The node index encodes each regulatory node using its code, title, text, notes, exclusions, and definitions. The evidence index encodes evidence chunks together with their document title and local parent-node context. Both node and evidence retrieval use cosine similarity between the query embedding and the corresponding indexed vectors.

\textbf{Query preparation and anchor initialization.} For each input instance, the LLM first extracts a retrieval-oriented query, anchor terms, and constraint terms from the original description. These terms preserve the product entity and rule-relevant attributes while reducing irrelevant wording. In the default setting, the first search step uses fast anchor initialization with $topM=8$ to construct a global candidate set from the full node index. This uses the same candidate-size parameter as local candidate retrieval, but applies it globally over all indexed nodes. The global candidates are processed by the same candidate package construction and next-hop decision procedure as local candidates. If this global retrieval step returns no candidate, the system falls back to expanding the children of the root node.

\textbf{Local candidate retrieval.} After the initial anchor step, search proceeds only through valid child nodes of the current node. At each depth, candidate children are ranked by cosine similarity between the query embedding and node embeddings. The local candidate size is controlled by $topM$; in the main HS experiments, we use $topM=8$ unless otherwise specified. For each candidate node, the system retrieves one top-ranked evidence chunk for branch selection. The full candidate package contains code, title, node type, retrieval score, node text, notes, exclusions, definitions, and retrieved evidence.

\textbf{LLM branch selection.} The LLM receives the current query, the current node, and the candidate packages. It selects the next node, provides a confidence score and rationale, and may return a stop signal when the current candidate is sufficiently specific. The output is parsed into a selected child identifier, stop flag, confidence score, and rationale. For the candidate-field ablation, we vary the visible fields from title-only to the full candidate package. The default setting uses the full package.

\textbf{Stopping and evidence aggregation.} Each single-pass search stops when the LLM returns a stop signal, the selected node has no children, or the maximum depth is reached. We set the maximum depth to 8. For the final decision, evidence from the selected path is deduplicated and ranked. We keep up to two evidence chunks per visited node, at most six chunks per pass, and at most ten aggregated evidence chunks across rounds before invoking the final decision generator, which outputs the final class label together with confidence, evidence, and rationale.

\section{Baseline Implementation Details}
\label{app:baselines}

This appendix provides implementation details for the retrieval-augmented baselines used in our experiments. Unless otherwise specified, all baselines are evaluated on the same regulatory corpus, test split, embedding backend, and LLM as our method. Each baseline receives the same natural language input instance and is required to output a final regulatory code. The baseline parameters are kept fixed across datasets and are not tuned on the test set. For methods that require retrieved textual context, duplicate text chunks are removed before the final context is passed to the LLM.

\textbf{LightRAG.} We evaluate LightRAG in \texttt{hybrid} retrieval mode. This mode combines low-level keyword search over entity or node representations with high-level keyword search over relation or edge representations. The two result streams are merged in a round-robin manner to form the final context, allowing the model to use both precise entity matching and broader relational evidence. We set \texttt{top\_k}=40, which controls the maximum number of retrieved entities in local retrieval and the maximum number of retrieved relations in global retrieval. We set \texttt{chunk\_top\_k}=20, so that at most 20 deduplicated and reranked text chunks are included in the LLM context.

\textbf{Flat Evidence RAG.} The flat RAG baseline retrieves evidence chunks directly from the full regulatory corpus without using hierarchical constraints. We set \texttt{chunk\_top\_k}=20, so that at most 20 retrieved text chunks are retained after deduplication and reranking. This baseline tests whether global evidence retrieval alone is sufficient for regulation-driven fine-grained classification.

\textbf{GraphRAG.} For GraphRAG, we retrieve up to 10 entities and up to 10 relationships for each entity. We set \texttt{top\_k\_entities}=10 and \texttt{top\_k\_relationships}=10. In constructing the final context, we set \texttt{text\_unit\_prop}=0.5 and \texttt{community\_prop}=0.15, which control the relative contribution of original text units and community reports, respectively. These settings allow GraphRAG to combine local entity-level evidence with global community-level summaries.

\textbf{LinearRAG.} For LinearRAG, we set \texttt{top\_k\_sentence}=3, meaning that each entity selects the top three bridge sentences during each iteration to discover new entities according to cosine-similarity ranking. We set \texttt{max\_iterations}=3, allowing at most two propagation steps beyond the seed entities. Entity extraction is performed using the spaCy \texttt{zh\_core\_web\_sm} NER model, which is used to identify candidate entities from Chinese regulatory text.

\textbf{MA-RAG.} For MA-RAG, we set \texttt{top\_k}=2 for document retrieval at each retrieval step. The multi-agent pipeline consists of six sub-agents: \texttt{plan}, \texttt{step\_define}, \texttt{generate}, \texttt{extract}, \texttt{aggregate}, and \texttt{summary}. We set the temperature to 0.3 for the generative agents, including \texttt{plan}, \texttt{step\_define}, and \texttt{generate}, and to 0.0 for the deterministic extraction and aggregation agents, including \texttt{extract}, \texttt{aggregate}, and \texttt{summary}.

\printcredits

\bibliographystyle{cas-model2-names}

\bibliography{cas-refs}





\end{document}